\documentclass[sn-mathphys,Numbered]{sn-jnl}

\usepackage{graphicx}
\usepackage{multirow}
\usepackage{amsmath,amssymb,amsfonts}
\usepackage{amsthm}
\usepackage{mathrsfs}
\usepackage[title]{appendix}
\usepackage{textcomp}
\usepackage{manyfoot}
\usepackage{booktabs}
\usepackage{algorithm}
\usepackage{algorithmicx}
\usepackage{algpseudocode}
\usepackage{listings}
\usepackage[table]{xcolor}
\usepackage{lmodern}
\usepackage{array}
\usepackage{bbm}
\usepackage{stackengine}
\usepackage{subcaption}
\usepackage{arydshln}
\usepackage{float}
\setstackEOL{\\}

\definecolor{rocket1}{HTML}{E1BAA0}
\definecolor{rocket2}{HTML}{D2866B}
\definecolor{rocket3}{HTML}{BC5157}
\definecolor{rocket4}{HTML}{8E335B}
\definecolor{rocket5}{HTML}{5F2D51}
\definecolor{rocket6}{HTML}{5F2D51}
\definecolor{rocket7}{HTML}{301E37}

\begin{document}

\title[Article Title]{Multimodal Graph-based Classification of Esophageal Motility Disorders} 

\author*[1]{\fnm{Alexander} \sur{Geiger}}\email{alexander.geiger@tum.de}
\equalcont{These authors contributed equally to this work.}

\author*[1]{\fnm{Lars} \sur{Wagner}}\email{lars.wagner@tum.de}

\equalcont{These authors contributed equally to this work.}

\author[2,3]{\fnm{Daniel} \sur{Rueckert}}\email{daniel.rueckert@tum.de}

\author[4]{\fnm{Alois} \sur{Knoll}}\email{k@tum.de}

\author[1,5]{\fnm{Dirk} \sur{Wilhelm}}\email{dirk.wilhelm@tum.de}

\author[1,5]{\fnm{Alissa} \sur{Jell}}\email{alissa.jell@tum.de}

\affil[1]{\orgdiv{Technical University of Munich, School of Medicine and Health, TUM University Hospital Rechts der Isar}, \orgname{Research Group MITI}, \orgaddress{\city{Munich}, \country{Germany}}}

\affil[2]{\orgdiv{Technical University of Munich, School of Computation, Information and Technology}, \orgname{Chair for AI in Healthcare and Medicine}, \orgaddress{\city{Munich}, \country{Germany}}}

\affil[3]{\orgdiv{Imperial College London}, \orgname{Department of Computing}, \orgaddress{\city{London}, \country{UK}}}

\affil[4]{\orgname{Technical University of Munich, School of Computation, Information and Technology}, \orgdiv{Chair of Robotics, Artificial Intelligence and Real-Time Systems}, \orgaddress{\city{Munich}, \country{Germany}}}

\affil[5]{\orgdiv{Technical University of Munich, School of Medicine and Health, TUM University Hospital Rechts der Isar}, \orgname{Department of Surgery}, \orgaddress{\city{Munich}, \country{Germany}}}

\abstract{\textbf{Purpose:} Diagnosing esophageal motility disorders, including dysphagia, pose significant challenges due to the complexity of high-resolution impedance manometry (HRIM) data and variability in clinical interpretation. This work explores the feasibility of a multimodal Machine Learning (ML)-based classification approach that combines HRIM recordings with patient-specific information and incorporates a graph-based modeling of esophageal physiology.

\textbf{Methods:} We analyze HRIM recordings with corresponding patient information from 104 patients with esophageal motility disorders collected at TUM University Hospital. Patient data includes demographic, clinical, and symptom information extracted from structured questionnaires and free-text notes using keyword detection and large language model-based processing. HRIM data is represented as spatio-temporal graphs, where nodes correspond to pressure values along the esophagus and edges encode spatial adjacency and impedance dynamics. A graph neural network (GNN) is applied to learn physiologically meaningful representations, which are fused with patient embeddings for multi-category, multi-class classification of swallow events. The impact of patient features and graph-based modeling is evaluated by ablation studies and comparison to vision-based classifier baselines.

\textbf{Results:} The proposed multimodal approach, incorporating patient-specific information, indicates improvements over models that rely solely on HRIM-derived features across all classification categories. Additionally, the graph-based modeling provides gains compared to vision-based baselines. Our experiments systematically assess the complementary contribution of multiple modalities, as well as demonstrate the feasibility of our proposed graph-based approach.

\textbf{Conclusion:} Our initial findings demonstrate that integrating patient-level data with graph-based representations of HRIM signals appears to be a promising direction for more accurate classification of esophageal motility disorders. For further validation, future studies should include larger and more representative datasets to confirm these trends and ensure generalizability.
}


\maketitle

\section{Introduction}\label{introduction}

Benign esophageal diseases pose substantial health and socio‑economic burdens, particularly in aging populations. Dysphagia, defined as difficulty swallowing food, liquids, or even saliva, becomes increasingly common with advancing age and presents significant challenges for both patients and healthcare systems~\cite{hunterDysphagiaAgingPopulation2024, smithTrueCostDysphagia2023, allenEconomicCostsDysphagia2020}. These disorders may originate from impairments across the oral, pharyngeal, or esophageal phases of swallowing, and their intermittent nature often makes them difficult to diagnose~\cite{sheehanDysphagiaOtherManifestations2008, wilkinsonEsophagealMotilityDisorders2020}.
High-resolution manometry (HRM) is the current gold standard for evaluating esophageal motility disorders~\cite{bredenoordHighresolutionManometry2008, carlsonHighResolutionManometryClinical2015}
During HRM, a transnasally inserted catheter equipped with multiple pressure sensors records intraluminal esophageal motility patterns along its length.
Additionally, HRM can also be combined with impedance measurement using a specialized catheter, a technique referred to as high-resolution impedance manometry (HRIM), which allows simultaneous assessment of bolus transit and pressure dynamics.
Interpreting HRIM recordings is inherently challenging. The procedure generates complex spatio-temporal pressure and impedance patterns that require substantial expertise to analyze accurately. Clinicians must undergo extensive training to recognize subtle deviations from normal motility, and even experienced practitioners can face difficulties when distinguishing between overlapping or borderline patterns. This results in a limited inter-rater reliability of examination results due to varying expertise of clinicians~\cite{foxInterobserverAgreementDiagnostic2015, kimFactorsDeterminingInterobserver2018}. These challenges contribute to variability in interpretation and underscore the need for more robust, standardized diagnostic approaches.

To address these limitations, prior research has primarily focused on applying ML-based methods to analyze HRIM data,  aiming to improve classification accuracy, reduce inter-observer variability, and support clinical decision-making. However, existing approaches rely almost exclusively on HRIM-derived features and do not incorporate patient-specific information, such as demographic, clinical, or comorbidity data. Integrating these factors could enable more personalized and comprehensive assessments. Therefore, in this work we explore a multimodal approach to classify esophageal motility disorders, combining patient information with HRIM.

Furthermore, while it has been investigated to use a graph-based approach in analyzing HRM data, no work exists that directly models the anatomical and physiological structure of the esophagus as a graph. Therefore, we additionally propose a method that represents HRIM data itself as a graph, where nodes correspond to pressure sensors and edges encode spatial adjacency and impedance dynamics. This design enables graph-based learning to capture intrinsic topological and physiological dependencies of esophageal motility without depending on handcrafted feature engineering, aiming to improve diagnostic accuracy.

\begin{figure}[t]
\centering
\includegraphics[width=\linewidth]{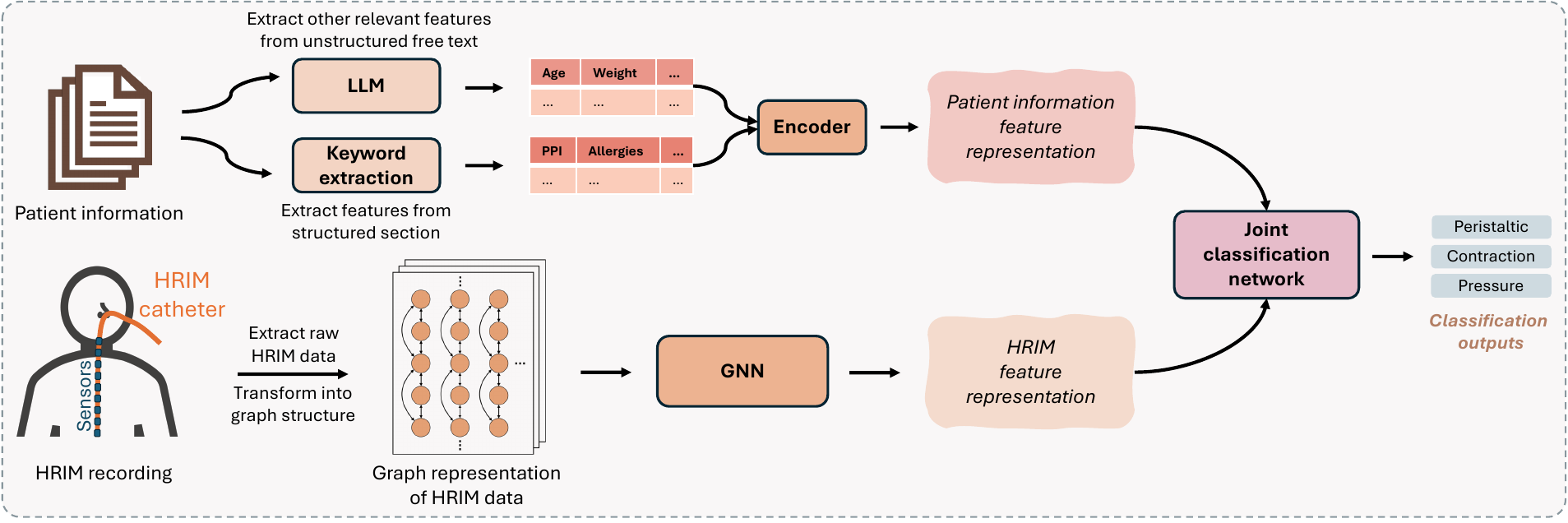}
\caption{Overview of our proposed multimodal, graph-based classification pipeline.}
\label{fig:visual_abstract}
\end{figure}

The overview of our proposed multimodal, graph-based classification pipeline can be seen in Fig.~\ref{fig:visual_abstract}.

\section{Motivation and related work}\label{motivation}

Computer-aided approaches, including artificial intelligence (AI), have been widely explored to improve the analysis and diagnosis of esophageal disorders~\cite{gongAIEsophagealMotility2025}.
Most efforts focus on HRM and impedance data, addressing tasks such as sphincter motility analysis~\cite{jungheimCalculationUpperEsophageal2016, leeHighresolutionManometryReliability2014}, 
swallow detection in long-term HRM~\cite{jellHowCopeBig2020, geigerDeepLearningbasedApproach2025}
, and visualization~\cite{geigerMultimodalVisualizationEsophageal2025}.
In addition, automated swallow assessment has also been actively studied. In this domain, early work applied simple neural networks on HRM data for pattern recognition and classification into specified categories.~\cite{hoffmanArtificialNeuralNetwork2013, hoffmanClassificationHighresolutionManometry2013}. Other approaches include physiological parametric models for swallow categorization~\cite{carnielPhysiologicalModelInvestigation2016, frigoProcedureAutomaticAnalysis2018} and using distension-contraction features for motility disorder classification~\cite{zifanUnravelingFunctionalDysphagia2023, zifanEnhancingDiagnosticYield2024}.
Recent work increasingly leverages deep learning, such as image-based classification approaches with CNNs~\cite{popaAutomaticDiagnosisHighResolution2022, surdea-blagaAutomatedChicagoClassification2022}, LSTM-based swallow classification and multi-stage pipelines~\cite{kouDeepLearningBased2022, kouMultistageMachineLearning2022}, and CNN–BiLSTM hybrids for sequence analysis~\cite{wangDeepLearningTracing2021}.
Other approaches include graph-based feature selection with fuzzy classification~\cite{rafieivandFuzzybasedFrameworkDiagnosing2023} and ensemble learning with attention mechanisms~\cite{wuMixedAttentionEnsemble2025}.
These methods report classification accuracies of up to 98\%, though most rely on internal datasets and address varying classification targets, limiting comparability.

Despite these advances, prior work has focused almost exclusively on HRIM-derived features. To our knowledge, no study has integrated additional patient-specific information, such as demographic or clinical data, into the modeling process so far, even though associations between patient characteristics and swallowing disorders have been investigated. Previous research has explored relationships involving age~\cite{shimEffectsAgeEsophageal2017, kunenEsophagealMotilityPatterns2020}, gender~\cite{dantasGenderEffectsEsophageal1998, kamalGenderMedicationUse2018}, heartburn~\cite{takahashiClinicalCharacteristicsEsophageal2021}, obesity~\cite{leNormativeHighResolution2024}, medication use, comorbidities, and other clinical characteristics~\cite{kamalGenderMedicationUse2018}, as well as ethnic differences~\cite{cohenEthnicDifferencesClinical2023}. These findings underscore the potential value of incorporating patient-specific factors into computational models to enable more comprehensive and personalized assessments of esophageal motility disorders.

In addition, while graph-based approaches have been explored in this domain, they have not directly modeled the anatomical and physiological structure of the esophagus. For example,~\citet{rafieivandFuzzybasedFrameworkDiagnosing2023} constructed a feature-level graph after computing correlations and extracting spatio-temporal features from HRM signals, using this representation for feature selection prior to fuzzy classification. In contrast, our approach represents the HRIM data itself as a graph, where nodes correspond to catheter sensors and encode pressure values, while edges encode spatial adjacency and impedance values, exploiting the intrinsic topology and physiological dependencies without relying on correlation-based feature engineering.

\section{Methods}\label{method}

\subsection{Data set}
\begin{figure}[!t]
\centering
\includegraphics[width=0.98\linewidth]{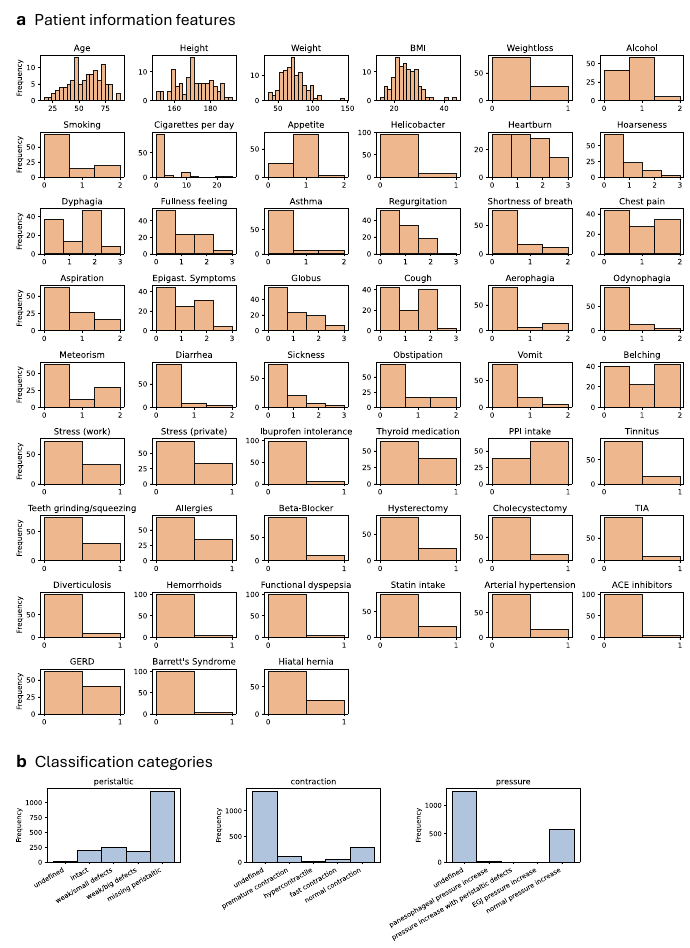}
\caption{(a)~The patient information features with their respective distributions. (b)~The target classification categories with their respective distributions.}
\label{fig:distributions}
\end{figure}

The dataset used in this study was collected at the TUM University Hospital in Munich and comprises HRIM examinations from 104 patients with esophageal motility disorders between 2020 and 2025. Prior to the procedure, each patient completed a standardized questionnaire together with the attending medical personnel. This document includes structured sections capturing demographic information, comorbidities, and symptom profiles, as well as a free-text field for additional clinical notes provided by the examiner.

\begin{figure}[t]
\centering
\includegraphics[width=\linewidth]{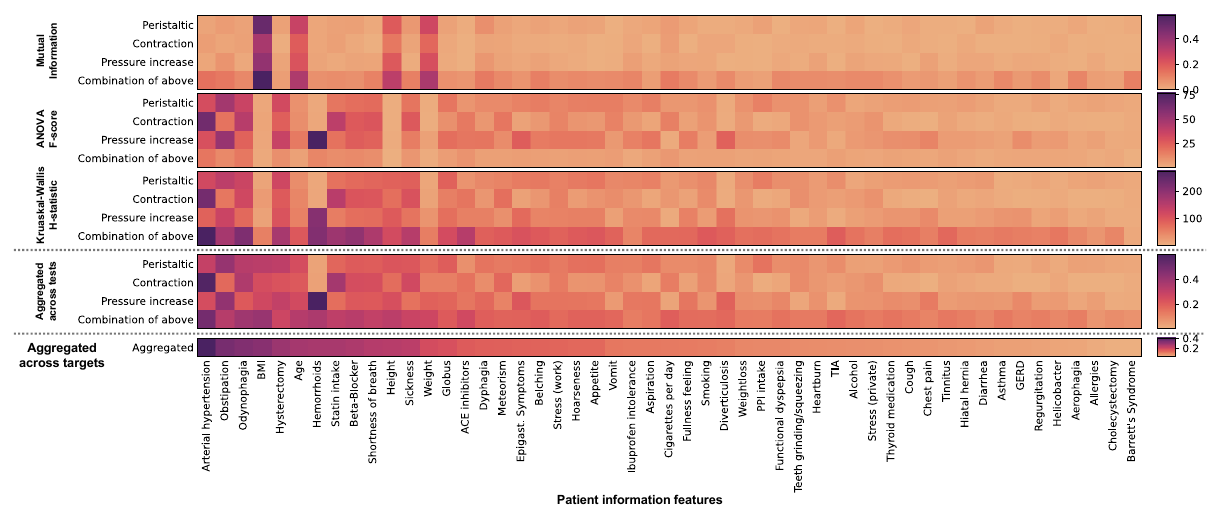}
\caption{Correlations of patient information and targets, with aggregations across measures and target categories.}
\label{fig:correlations}
\end{figure}

HRIM examinations were conducted according to the Chicago Classification protocol version 3.0~\cite{kahrilasChicagoClassificationEsophageal2015}. For each swallow event (in total $\sim$1,800 swallows across all patients), medical personnel annotated three distinct categories: peristaltic pattern, contraction type, and pressure characteristics. Each category contains multiple possible labels, resulting in a multi-category, multi-class classification task for our predictive model. Fig.~\ref{fig:distributions}~(b) shows the target categories and their respective values and distribution, reflecting a strong class imbalance due to the natural prevalence of certain patterns.

\subsection{Patient information feature representation}\label{sec:patient_info}
Information from the standardized questionnaire was processed using two complementary approaches. First, structured fields were extracted via keyword-based detection, yielding 4 numeric features (e.g., age, BMI) and 28 categorical or ordinal features (e.g., weight loss, smoking).
Second, unstructured clinical notes from the free-text field were analyzed using a large language model (LLM) (in our experiments Qwen3-14B~\cite{qwen3technicalreport}). The LLM was initially provided with all free-text entries to generate a comprehensive list of clinically relevant attributes. Subsequently, the LLM processed each individual note to identify the presence or absence of these attributes, resulting in 19 binary features.
Thus, 51 patient-specific features were obtained in total. Fig.~\ref{fig:distributions}~(a) shows the features as well as their distributions across the patients which are used in our evaluation. All features were standardized to zero mean and unit variance. For a single patient, this results in a patient information feature vector \(\mathbf{x}^{(\text{p})} \in \mathbb{R}^{1 \times 51}\). These features are then encoded using a simple feed-forward encoder neural network, producing a dense representation suitable for integration with HRIM-derived features in the multimodal classification pipeline.

For an initial assessment of the relationship between patient features and target classes, we compute three correlation measures for each feature against the three individual classes and the combined target. Each correlation score is scaled to [0,1] and averaged across the three measures to obtain a single value per target category. Finally, we average these values across targets, yielding one representative correlation score for each feature. The correlation results are shown in Fig.~\ref{fig:correlations}.

\subsection{Graph-based HRIM feature representations}
The HRIM data is represented as a spatio-temporal graph to capture the anatomical and functional organization of the esophagus in a domain-specific way. The esophagus is a sequential structure and the HRIM catheter includes multiple pressure and impedance sensors along its length, recording bolus transit and pressure wave propagation. By modeling pressure sensors as nodes and their spatial relationships and impedance values as edges, we encode the medical structure of the esophagus directly into the graph. This enables graph-based learning, which is well-suited for capturing complex dependencies across space and time. Such a representation provides a physiologically meaningful alternative to traditional feature-based approaches.

\begin{figure}[t]
\centering
\includegraphics[width=\linewidth]{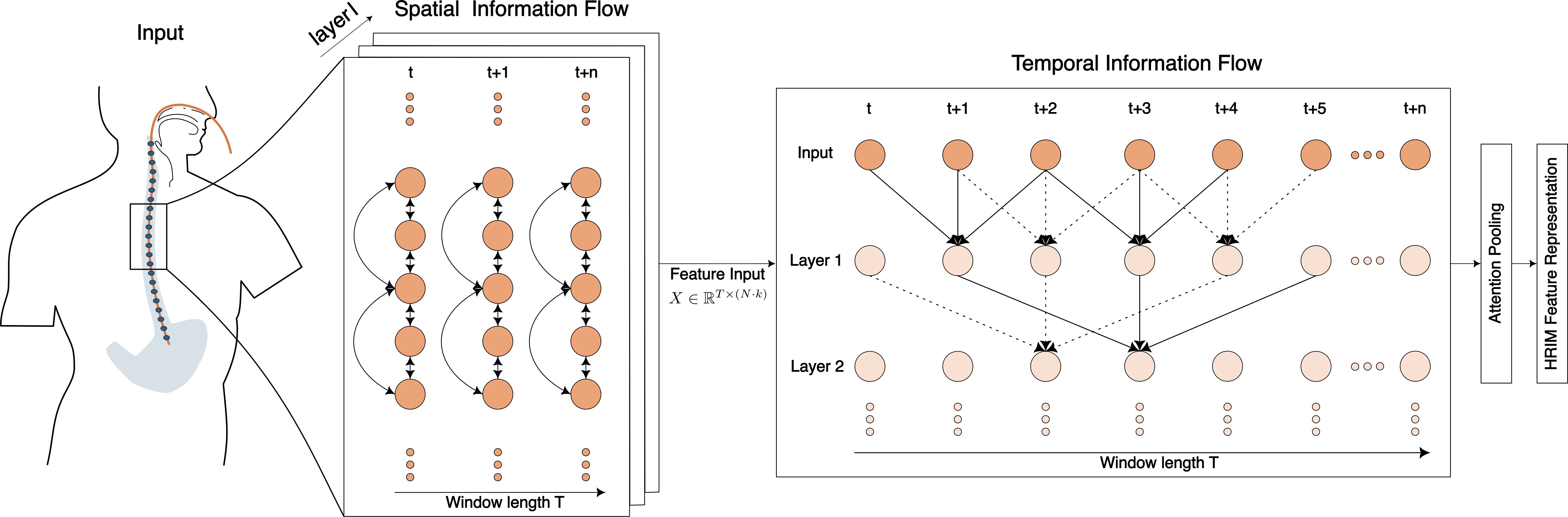}
\caption{Overview of the proposed spatio-temporal graph-based HRIM modeling framework. 
}
\label{fig:graph_setup}
\end{figure}

The overview of the graph-based feature representation can be seen in Fig.~\ref{fig:graph_setup}. 

The HRIM data originates from 36 pressure sensors and 15 impedance sensors along the catheter. For each swallow, we use 750 time steps (15 seconds), forming spatio-temporal manometry and impedance matrices \( \mathbf{X}^{(\text{m})} \in \mathbb{R}^{750 \times 36}\) and \( \mathbf{X}^{(\text{i})} \in \mathbb{R}^{750 \times 15}\).
To represent this information as a graph, we model a swallow event as a fixed-length sequence of 750 graphs $\mathcal{G}= \{ \mathcal{G}_{1}, ..., \mathcal{G}_{750} \}$, where each time step $t$ of a swallow event is modeled as a graph $\mathcal{G}_t = (\mathcal{V}_t, \mathcal{E}_t)$, with  a set of nodes $\mathcal{V}_t$ and a set of edges $\mathcal{E}_t$. Each node $v_i \in \mathcal{V}_t$ represents a pressure sensor $i \in \{1, 2, \dots, 36\} $ along the catheter. The feature of each node at time step $t$, $\mathbf{h}_i^t \in \mathbb{R}^d$, consists of the pressure value recorded by sensor \( i \) at time $t$. The nodes are connected along the catheter via edges $e_{ij} \in \mathcal{E}_t$, representing the spatial adjacency between neighboring sensors. Every second node, beginning from the third sensor, is connected via an additional edge reflecting the impedance measurements of the catheter. This impedance relationship is encoded as an edge feature $\mathbf{f}_{ij}^t \in \mathbb{R}^d$, capturing the physiological interaction between sensor pairs at each time step $t$.\par
We apply a GNN layer to aggregate spatial information, updating the features of a node $h_i^{(l)}$ at layer $l$ by aggregating messages from its neighboring nodes $j \in N(i)$:
\begin{equation}
h_i^{(l)} = \sigma \; AGG^{(l)}\left(\left\{\;MSG^{(l)} \left(\left[\mathbf{h}_{j}^{(l-1)} \big\Vert \mathbf{f}_{ij}^{(l-1)}\right]+b \right), \forall j \in N(i)\right\}\right).
\label{eq:a}
\end{equation}
Here, the node features $\mathbf{h}_j^{(l-1)}$ (pressure) and edge features $\mathbf{f}_{ij}^{(l-1)}$ (impedance) are concatenated during message passing. The functions $MSG$ and $AGG$ are defined according to the specific GNN layer used (in this work: GATv2~\cite{brodyHowAttentiveAre2022} and GENConv~\cite{liDeeperGCNAllYou2020}).\par
For temporal modeling, the graph data at each time step $t$ is represented as a feature matrix $X_t \in \mathbb{R}^k$, where $k$ corresponds to the number of features per node. The sequence of these feature matrices $\{ X_t \}_{t=1}^{750}$ is then passed into a temporal model (in this work a Temporal Convolutional Network (TCN) and a Transformer), producing a compact embedding of the swallow event. This embedding is followed by hierarchical attention pooling. First, node-level attention pooling aggregates information across sensors at each time step, yielding a compact representation of the spatial esophageal activity. Subsequently, temporal attention pooling summarizes the evolution of this representation over the entire swallow sequence. This pooling strategy is performed independently for all three classification categories (peristaltic pattern, contraction type, and pressure characteristics), allowing the model to learn category-specific representations.

\subsection{Multimodal classification}

In the joint classification network, the graph-based category-specific representations of a swallow event are fused with the patient information feature representation using a fusion module, which employs a simple concatenation of both embeddings. Then, each fused embedding is passed through a category-specific multilayer perceptron to produce the corresponding class predictions.

The model is trained in a multi-category classification setting with three parallel classification heads, one per category. Each category \(c \in \{1,2,3\}\) predicts a label from a category-specific label set \(\mathcal{Y}^{(c)}\). To mitigate class imbalance, we employ a class-weighted cross-entropy loss with label smoothing for each category. For a sample with ground-truth label \(y_i^{(c)} \in \mathcal{Y}^{(c)}\), we compute the class-weighted cross-entropy loss with label smoothing as
\begin{equation}
	\mathcal{L}_{\mathrm{ce}}^{(c)} 
	= - \frac{1}{N} \sum_{i=1}^{N} 
	\sum_{k \in \mathcal{Y}^{(c)}} 
	w_k^{(c)} \, \tilde{y}_{i,k}^{(c)} \, \log p_{i,k}^{(c)},
\end{equation}
where \(N\) is the batch size, \(w_k^{(c)}\) is the weight associated with class \(k\) for category \(c\) (computed as an inverse function of the training fold class frequencies), and \(p_{i,k}^{(c)}\) denotes the predicted probability for class \(k\). The smoothed target distribution \(\tilde{\mathbf{y}}_i^{(c)}\) is given by
\begin{equation}
	\tilde{y}_{i,k}^{(c)} =
	(1 - \varepsilon)\,\mathbb{I}[k = y_i^{(c)}] 
	+ \frac{\varepsilon}{|\mathcal{Y}^{(c)}|},
\end{equation}
with label smoothing parameter \(\varepsilon \in [0,1)\) and indicator function \(\mathbb{I}[\cdot]\).

\begin{figure}[t]
     \centering
     \begin{subfigure}{0.32\textwidth}
         \centering
         \includegraphics[width=\linewidth]{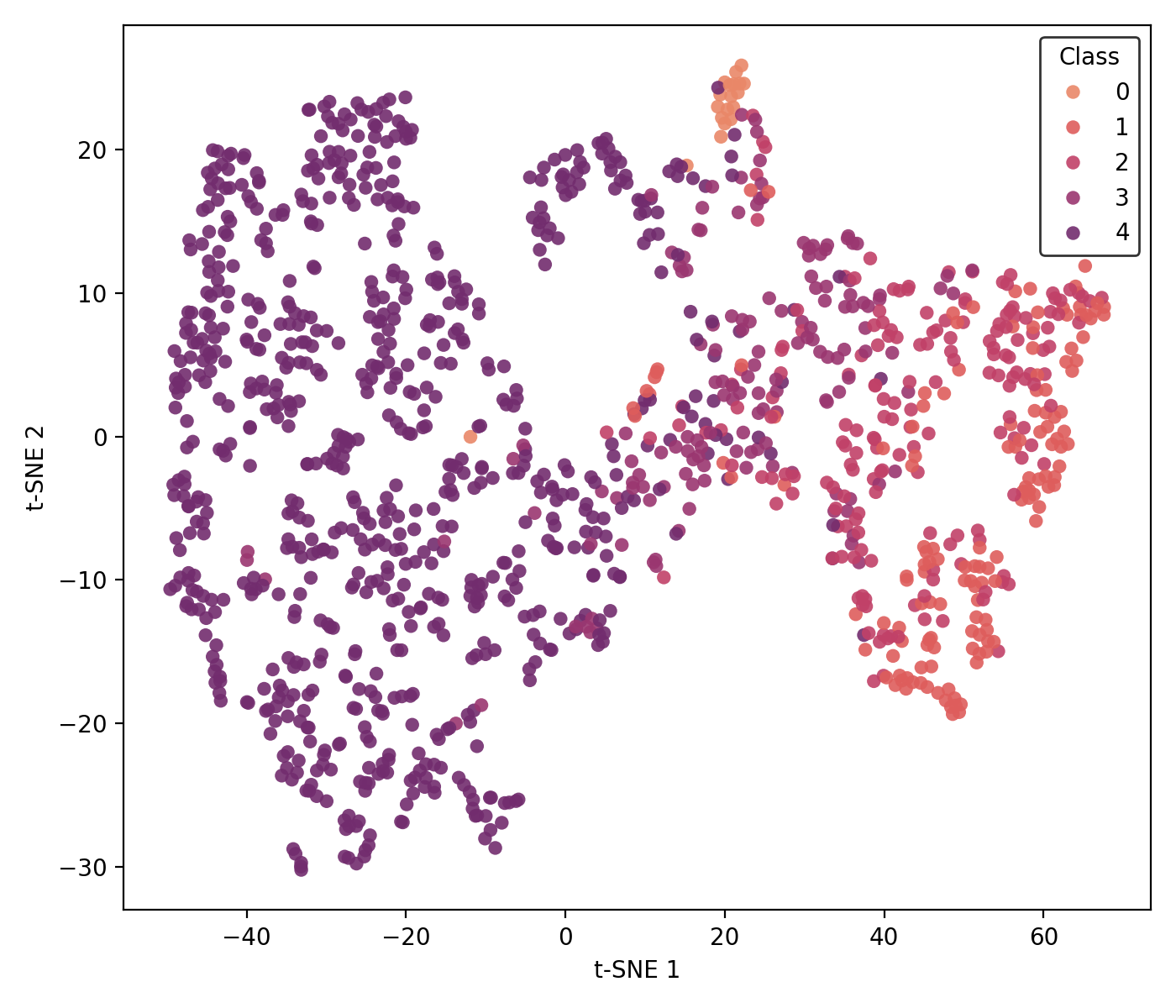}
         \caption{peristaltic pattern}
         \label{fig:tsne_class1}
     \end{subfigure}
     \hspace{0.01cm}
     \begin{subfigure}{0.32\textwidth}
         \centering
         \includegraphics[width=\linewidth]{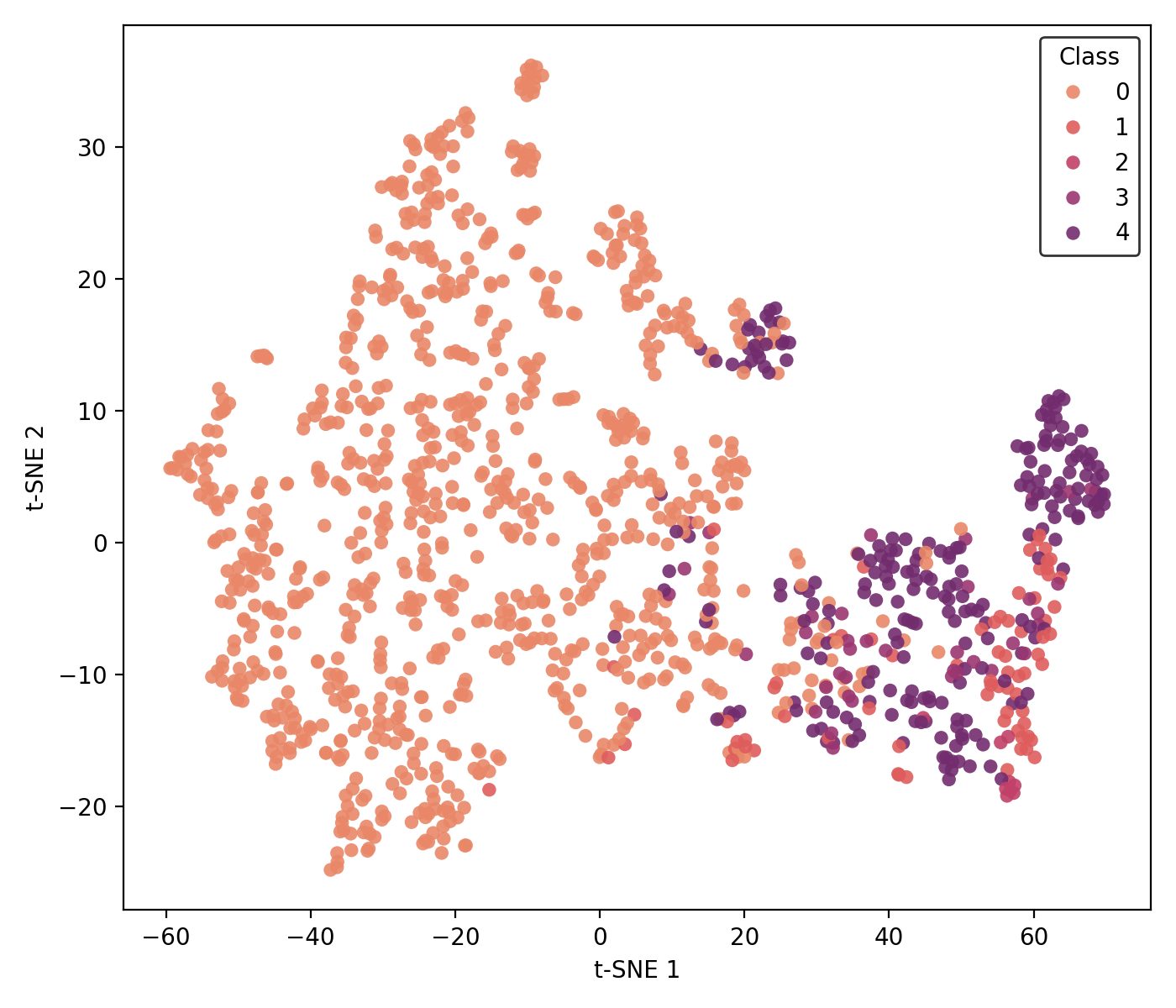}
         \caption{contraction type}
         \label{fig:tsne_class2}
     \end{subfigure}
      \hspace{0.01cm}
     \begin{subfigure}{0.32\textwidth}
         \centering
         \includegraphics[width=\linewidth]{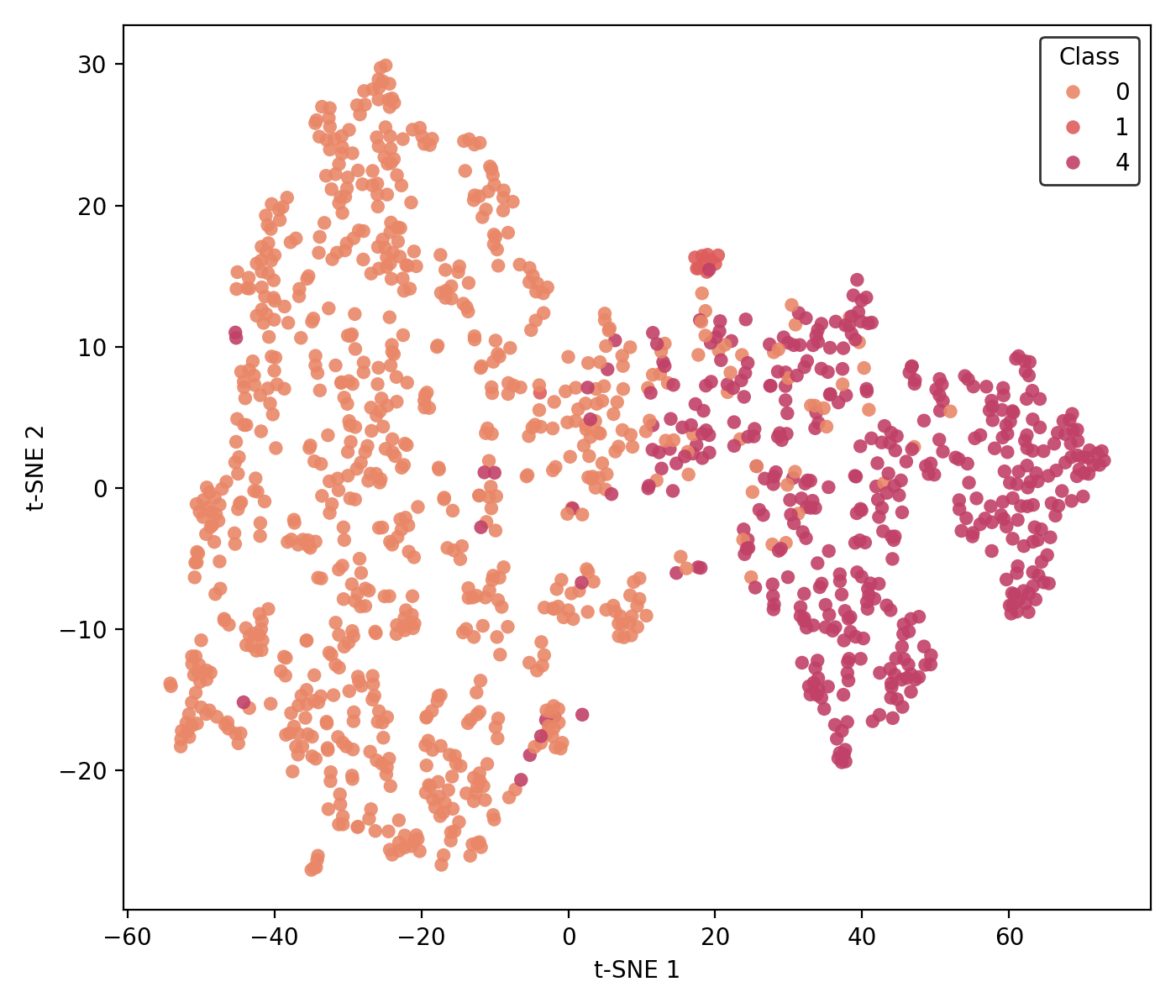}
         \caption{pressure characteristics}
         \label{fig:tsne_class3}
     \end{subfigure}
     \caption{t-SNE visualization of post-training embeddings of swallows events for each classification target projected into two dimensions, colored by class label.}
     \label{fig:tsne}
\end{figure}

To further regularize the learned representations, we apply a supervised contrastive loss~\cite{khoslaSupervisedContrastiveLearning2020} to the category-specific embeddings, encouraging samples sharing the same class label to form compact clusters in the embedding space.
For category \(c\), the supervised contrastive loss is defined as
\begin{equation}
	\mathcal{L}_{\mathrm{con}}^{(c)} 
	= \frac{1}{N} \sum_{i=1}^{N} 
	\frac{-1}{|\mathcal{P}_i^{(c)}|}
	\sum_{p \in \mathcal{P}_i^{(c)}}
	\log
	\frac{
		\exp\left( \mathrm{sim}(\mathbf{z}_i^{(c)}, \mathbf{z}_p^{(c)}) / \tau \right)
	}{
		\sum\limits_{\substack{a=1 \\ a \neq i}}^{N}
		\exp\left( \mathrm{sim}(\mathbf{z}_i^{(c)}, \mathbf{z}_a^{(c)}) / \tau \right)
	},
\end{equation}
where \(\mathbf{z}_i^{(c)} \in \mathbb{R}^{d_c}\) denotes the category-specific embedding of sample \(i\) for category \(c\), \(\mathcal{P}_i^{(c)} = \{ p \neq i \mid y_p^{(c)} = y_i^{(c)} \}\) denotes the set of indices corresponding to samples that share the same label as \(i\), \(\tau > 0\) is a temperature hyperparameter, and \(\mathrm{sim}(\cdot,\cdot)\) denotes the cosine similarity between embeddings. Figure~\ref{fig:tsne} illustrates the post-training embeddings projected into two dimensions using t-SNE, with points colored by class label, demonstrating that the learned representations exhibit clear inter-class separation.

The overall training objective combines both losses across all three classification categories:
\begin{equation}
	\mathcal{L} 
	= \sum_{c=1}^{3} \mathcal{L}_{\mathrm{ce}}^{(c)} 
	+ \lambda_{\mathrm{con}} \sum_{c=1}^{3} \mathcal{L}_{\mathrm{con}}^{(c)},
\end{equation}
where \(\lambda_{\mathrm{con}}\) is a scalar hyperparameter controlling the relative contribution of the supervised contrastive regularization term.

\section{Results}

\subsection{Experimental setup}

We evaluate our approach using a 5-fold cross-validation defined at the patient level to avoid information leakage between training and validation sets. To construct balanced folds in the presence of multiple target variables, we employ an iterative multilabel stratified k-fold strategy. For each patient, a fixed-length label representation is computed by aggregating all swallow-level labels into normalized histograms for each classification category. These patient-level label vectors are then used for stratified splitting, ensuring that the joint label distribution across all three categories is preserved across folds.\par
Model performance is evaluated using weighted averaged F1 scores (wAF1) for each classification category to account for class imbalance. All experiments were implemented in PyTorch and performed on a single NVIDIA L40S GPU.

\subsection{General results}

\begin{table}[t]
\centering
\small
\caption{Classification results comparing vision- and graph-based models. Mean weighted F1-score ($\pm$ standard deviation) over five-fold cross-validation is shown for the three classification categories. The two best values across category are highlighted.}
\begin{tabular}{
p{0.05\linewidth}
p{0.22\linewidth}
>{\centering\arraybackslash}p{0.17\linewidth}
>{\centering\arraybackslash}p{0.17\linewidth}
>{\centering\arraybackslash}p{0.17\linewidth}
}
    \toprule
    & Model & $wAF1_{peristaltic}$ & $wAF1_{contraction}$ & $wAF1_{pressure}$ \\
    \midrule
    \multirow{3}{*}{\rotatebox[origin=c]{90}{\textbf{vision}}} & 
    ResNet50            & $74.82 \pm 7.28$ & \cellcolor{rocket2!30}$80.94 \pm 1.70$ & $86.61 \pm 3.83$ \\
    & ViT               & $74.21 \pm 4.99$ & $79.93 \pm 2.66$ & $87.18 \pm 1.88$ \\
    & Swinv2            & $64.43 \pm 17.78$ & $74.43 \pm 12.96$ & $76.23 \pm 19.30$ \\
    \hdashline[2pt/2pt]
\noalign{\vskip 0.3ex}
    \multirow{4}{*}{\rotatebox[origin=c]{90}{\textbf{graph}}} & 
    Gatv2-TCN           & $63.88 \pm 6.84$ & $75.71 \pm 6.74$ & $80.44 \pm 8.46$ \\
    & Gatv2-Transformer & $66.30 \pm 4.30$ & $77.73 \pm 3.77$ & $83.78 \pm 4.95$ \\
    & Gen-TCN           & \cellcolor{rocket2!30}$74.98 \pm 4.34$ & \cellcolor{rocket3!40}$83.05 \pm 3.18$ & \cellcolor{rocket3!40}$88.47 \pm 1.56$ \\
    & Gen-Transformer   & \cellcolor{rocket3!40}$75.96 \pm 7.63$ & $80.78\pm 5.06$ & \cellcolor{rocket2!30}$87.44 \pm 2.44$ \\
    \bottomrule
\end{tabular}
\label{tab:general_results}
\end{table}
We compare the graph-based HRIM models with three vision-based baselines (i.e. ResNet50~\cite{heDeepResidualLearning2016}, Vision Transformer (ViT)~\cite{dosovitskiyImageWorth16x162021}, and Swin Transformer V2 (Swinv2)~\cite{liuSwinTransformerV22022}) to assess the impact of different spatial–temporal modeling choices. For the vision-based baselines, each swallow event is transformed into an image-like representation by resampling and reshaping pressure and impedance data into a two-channel input tensor $X \in \mathbb{R}^{2\times224\times224}$. To obtain a three-channel representation compatible with standard visual backbones, a third channel is generated via an attention-based mixing of the two modalities.\par
Table~\ref{tab:general_results} summarizes the overall classification performance across all evaluated models, reported as weighted F1 scores for each classification category (peristaltic pattern, contraction type, and pressure characteristics). Results are averaged over the five folds and reported as mean $\pm$ standard deviation.\par
Across all three classification category, the proposed graph-based approaches (especially those using GEN-Conv) consistently achieve the highest mean performance, with the Gen-TCN model showing the strongest overall results.
While statistical tests (Friedman test: peristaltic pattern $p=.151$, contraction type $p=.592$, pressure characteristics $p=.547$) and pairwise Wilcoxon signed-rank with Holm correction did not indicate significant differences between the models (likely due to limited number of cross-validation folds ($n=5$), the limited sample size and class imbalance), the performance patterns and noticeable mean differences consistently favor Gen-TCN. These results indicate a promising trend that warrants further investigation with larger, more representative cohorts.

\subsection{Ablation tests}
In order to assess the impact of each modality, we structurally ablate them, i.e. train and test the best model (Gen-TCN) without the respective modality. Table~\ref{tab:ablation_tests} reports the weighted averaged F1 scores for each output class. The model using all modalities achieved the highest overall performance across all three output classes, indicating the benefit of our multimodal integration. Removing individual modalities resulted in consistent performance degradation, with the largest drops observed when manometry was excluded, confirming expectations that manometry is the dominant modality. 
Removing patient information caused a smaller drop, suggesting that while it adds value, its contribution is less critical in our dataset. Interestingly, when comparing just manometry alone to manometry combined with either impedance or patient information, manometry by itself performed slightly better. This may indicate that impedance and patient features are conditionally informative but individually misleading or that a more sophisticated integration strategy is required to fully leverage their potential.

To also statistically assess the effect of modality ablation, we applied a Friedman test for each output class. Results indicated a significant global effect of data ablation for peristaltic pattern ($p=.0088$), contraction type ($p=.0106$), and pressure characteristics ($p=.0021$), confirming that input composition significantly influences model performance across all targets.
Post-hoc pairwise comparisons using Wilcoxon signed-rank tests with Holm correction did not reach significance, likely due to the same dataset related reasons listed above. Nevertheless, performance trends were consistent across classification categories. Configurations lacking certain modalities showed substantial mean reductions (approximately 0.13–0.20 in weighted F1), underscoring the critical role of choosing the right modalities in achieving robust performance.

\setlength{\tabcolsep}{0pt}
\begin{table}[t]
\centering
\small
\caption{Modality ablation results of the best model (Gen-TCN). Mean weighted F1-score ($\pm$ standard deviation) over five-fold cross-validation is shown for the three classification categories. Checkmark (\checkmark) denotes included modalities. The three best values across categories are highlighted.}
\begin{tabular}{
>{\centering\arraybackslash}p{0.135\linewidth}
>{\centering\arraybackslash}p{0.12\linewidth}
>{\centering\arraybackslash}p{0.105\linewidth}
@{\hspace{0.04\linewidth}} 
>{\centering\arraybackslash}p{0.18\linewidth}
>{\centering\arraybackslash}p{0.18\linewidth}
>{\centering\arraybackslash}p{0.18\linewidth}
}
    \toprule
    \multicolumn{3}{c}{\textbf{Modality}} & \multicolumn{3}{c}{\textbf{Metrics}} \\
    \cmidrule(l{0.3em}r{2.5em}){1-3} \cmidrule(l{0.3em}r){4-6}
    Manometry & Impedance & Patient & $wAF1_{peristaltic}$ & $wAF1_{contraction}$ & $wAF1_{pressure}$ \\
    \midrule
    \checkmark & \checkmark & \checkmark & \cellcolor{rocket3!40}$74.98 \pm 4.34$ & \cellcolor{rocket3!40}$83.05 \pm 3.18$ & \cellcolor{rocket3!40}$88.47 \pm 1.56$ \\
    \checkmark & \checkmark &            & $71.52 \pm 6.32$ & $79.68 \pm 7.22$ & \cellcolor{rocket2!30}$87.08 \pm 1.73$ \\
    \checkmark &            & \checkmark & \cellcolor{rocket1!20}$71.95 \pm 7.74$ & \cellcolor{rocket1!20}$79.81 \pm 6.86$ & $86.31 \pm 1.32$ \\
               & \checkmark & \checkmark & $57.12 \pm 10.04$ & $68.63 \pm 8.19$ & $72.37 \pm 12.14$ \\
    \checkmark &            &            & \cellcolor{rocket2!30}$74.73 \pm 3.46$ & \cellcolor{rocket2!30}$81.23 \pm 3.73$ & \cellcolor{rocket1!20}$86.74 \pm 3.09$ \\
               & \checkmark &            & $54.75 \pm 8.37$ & $65.92 \pm 7.23$ & $73.17 \pm 7.95$ \\
    \bottomrule
\end{tabular}
\label{tab:ablation_tests}
\end{table}

\section{Discussion and Conclusion}
The results indicate that the proposed graph-based approach achieves competitive performance and consistently outperforms vision-based baselines, although statistical significance could not be established with the current sample size. It is important to note that this work represents an initial exploration, and several methodological aspects, such as graph construction strategies, hyperparameter optimization, and integration of additional physiological features, offer opportunities for refinement.

A key limitation of this study is the relatively small dataset (104 patients) and the uneven distribution of classes. Expanding the dataset would improve robustness, enable more reliable statistical comparisons, and support generalization across diverse patient populations. Despite these constraints, the findings are encouraging: graph-based representations effectively capture the spatio-temporal dependencies inherent in HRIM data, and incorporating patient-specific information such as demographic and clinical features provides additional gains, underscoring the potential of multimodal modeling for personalized diagnostics.

In summary, this study provides initial evidence that combining graph-based modeling of HRIM signals with patient-level data can enhance automated classification of esophageal motility disorders. Future work should focus on scaling the dataset, refining graph architectures, and exploring advanced fusion strategies to further improve diagnostic accuracy and interpretability.

\section{Declarations}

\noindent\textbf{Funding}\par
\noindent No funding was received for conducting this study. \par

\noindent\textbf{Conflict of interest}\par
\noindent All authors (Alexander Geiger, Lars Wagner, Daniel Rueckert, Alois Knoll, Dirk Wilhelm, Alissa Jell) declare that they have no conflict of interest.\par

\noindent \textbf{Ethics approval}\par
\noindent All patient data was used anonymously with ethical approval and informed consent. Approval was granted by the local Ethics Committee of the TUM University Hospital rechts der Isar, Technical University Munich. The study was performed in line with the principles of the Declaration of Helsinki.

\bibliography{sn-bibliography}

\end{document}